RESEARCH PROJECT

**QUANTIFYING UNCERTAINTY IN RISK ASSESSMENT
USING FUZZY THEORY**

by

Hengameh Fakhravar

ENGINEERING MANAGEMENT AND SYSTEMS ENGINEERING

OLD DOMINION UNIVERSITY
April 2020

## ABSTRACT

Risk specialists are trying to understand risk better and use complex models for risk assessment, and many risks are not yet well understood. Some remain unknown, and new risks have arisen. Many risk types cannot be adequately analyzed using classical probability models. The lack of empirical data and complex causal and outcome relationships makes it difficult to estimate the degree to which certain risk types are exposed

Traditional risk models are based on classical set theory. While, fuzzy logic models are built on fuzzy set theory and fuzzy logic, and are useful for analyzing risks with insufficient knowledge or inaccurate data.

Fuzzy logic recognizes a lack of knowledge or a lack of accurate data, and it explicitly considers the cause and effect chain among variables. Most variables are described linguistically, in which fuzzy logic models most intuitively resemble human reasoning. These fuzzy models can help assess and learning about risks that are not well understood.

Fuzzy logic systems help to make large-scale risk management frameworks more simple. For risks that do not have an appropriate probability model, a fuzzy logic system can help model the cause and effect relationships, assess the level of risk exposure, rank key risks in a consistent way, and consider available data and experts' opinions. Besides, in fuzzy logic systems, some rules explicitly explain the connection, dependence, and relationships between model factors. This can help identify risk mitigation solutions. Resources can be used to mitigate risks with very high levels of exposure and relatively low hedging costs.

Fuzzy set and fuzzy logic models can be used with Bayesian and other types of method recognition and decision models, including artificial neural networks and decision tree models. These developed models have the potential to solve difficult risk assessment problems.

This research paper explores areas in which fuzzy logic models can be used to improve risk assessment and risk decision making. We will discuss the methodology, framework, and process of using fuzzy logic systems in risk assessment.

# TABLE OF CONTENTS

# 1   CHAPTER 1 - INTRODUCTION

Probabilistic risk models are prevalent in risk quantification and assessment. A classic probability model set theory may not be able to describe some risks realistically and functionally. Lack of familiarity with cause-and-effect relationships and imprecise data make it difficult to determine the degree of exposure to specific risk types using only typical probability models.

Risk assessment is a systematic process for identification, analysis, and evaluation. Identifying risk includes understanding the sources of risk, areas of influence, events, and their causes, and possible consequences that could positively or negatively affect the achievement of an enterprise's objectives. The aim is to create a detailed list of risks, including risks that may be associated with lost opportunities and risks out of the direct control of the organization. A comprehensive review allows full consideration of the potential effects of risk upon the organization.

The fuzzy logic and fuzzy set theory are invented by mathematician Lotfi A. Zadeh in 1965, to risk management. Unlike probability theory, fuzzy logic theory explicitly admits the uncertainty of truth; it also can easily incorporate information described in linguistic terms.

Fuzzy logic models are more convenient for integrating various expert opinions and more appropriate to case with insufficient and imprecise data. They include a system in which expert feedback and experience Dara will jointly evaluate uncertainty and recognize important issues.

The motivation for combining Risk Assessment and Fuzzy logic arises because Risk Assessment is often hindered by data limitations and ambiguity, such as incomplete or unreliable data, and subjective knowledge based on the reliance on human experts and their communication of linguistic variables. Since Fuzzy logic models are practical tools in such circumstances, it seems natural to inquire into the Risk Assessment applications of Fuzzy logic.(Tahami & Fakhravar, 2020a)

Interestingly, although widely accepted and sophisticated quantitative models can be used for assessing risks in various industries like credit and insurance risks, these risks are generally outside the control of managers. On the other hand, having sufficient risk detection and risk management in place can significantly reduce the functional risk, there is no global consensus on which quantitative model should be used. Thus, it may be more efficient to establish and implement better functional risk models utilizing newer methods such as fuzzy logic.

# 2    CHAPTER 2 - LITERATURE REVIEW

## 2.1    FUZZY BASED RISK MANAGEMENT METHODS

Since the first paper by (Zadeh, 1965) introduced the field of fuzzy theory, there has been a lot of academic research and practical implementation of fuzzy theory in almost all areas, from the physical to the social sciences. However, the focus here is to present the literature review related, directly or indirectly, to the field of risk management.

Several methods have already been used with the aim of developing risk management models. Project risks are often uncertain and vague in nature, leading to the use of the fuzzy concept.(Tahami et al., 2019) However, since fuzzy methods have several drawbacks, hybrid methods are increasingly being used. Generally, Fuzzy-based methods can be classified into three broad groups: (1) *basic fuzzy*, (2) *extended fuzzy,* and (3) *hybrid fuzzy* methods (Chan, 2009).

The *basic fuzzy* method can be defined as representing the basic concept of fuzzy logic and fuzzy set theory. The *extended fuzzy* method has modified algorithms based on fuzzy theory but not modified by other independent methods such as fuzzy arithmetic, fuzzy synthetic evaluation, fuzzy expert system, fuzzy Mamdani inference, fuzzy comprehensive evaluation, and fuzzy consensus qualitative analysis. The *hybrid fuzzy* method represents a combination of fuzzy and other independent methods. It involves different types, such as fuzzy probability methods, fuzzy matrix methods, fuzzy structured methods, the fuzzy cloud model, and fuzzy integral process. The following subsections briefly discuss fuzzy and hybrid methods most frequently in use.

### 2.1.1    BASIC FUZZY METHOD

#### 2.1.1.1    FUZZY LOGIC

Fuzzy logic has been used in risk evaluation for a long time, as it can be used to develop models on the basis of both qualitative data and quantitative values from historical records (Subramanyan et al., 2012), and is, therefore, a very effective management technique for achieving the objectives of projects under uncertainties, an impression (Abdul Rahman, 2013). Lyons and Skitmore (2004) and Novak (2012) described the common features of fuzzy logic as involving the following basic steps: (1) define and measure the likelihood of occurrence and severity of the risk in terms of verbal opinions and transform them into fuzzy numbers accordingly; (2) define a fuzzy inference to make a network between input and output parameters using fuzzy IF-THEN rules and/or "fuzzy arithmetic operators"; and (3) defuzzification the fuzzy outcomes into numerical values using appropriate quantifiers. Although existing fuzzy logic is a well-established theory, the lack of appropriate techniques to address fuzzy consistency and fuzzy priority vectors, together with the complicated operations involved, undermines its practical application (Zeng et al., 2007).

Fuzzy logic is in particular need of improvement for qualitative modeling data elicited from expert opinion using natural language. If there is a lack of knowledge about project risks, the qualitative data elicited from experts contains vagueness and uncertainty (Novák, 2006).

(Novák et al., 2012) discussed the potential of fuzzy mathematical logic to improve models affected by vagueness and suggested combining fuzzy logic with probability theory to capture the uncertainties involved. Further improvement of fuzzy logic is also necessary for modeling evaluative appropriate linguistic expressions of risk and developing suitable aggregation rules to quantify the linguistic expressions for risk ranking (Greaver et al., 2012; Novák et al., 2012).

### 2.1.1.2 FUZZY SET THEORY

Fuzzy set theory (FST) is a well-known decision support tool that allows experts to handle the uncertainty of events in risk assessment based on a linguistic assessment approach (Malek et al., 2015) This is an extended form of classical binary logic where a problem can only be considered as a full or non-member (i.e., 0 or 1). In practice, risks are generally not defined in this way due to the complexity and uncertainty of the problems (Chan, 2009). Contrary to binary logic, fuzziness is necessary to describe the gradual shift of an element from its membership in a set to a non-member state. Therefore, FST modifies the underlying binary logic to capture uncertainty and ambiguity in defining risk. For example, it is easy to define risk in linguistic terms such as extremely high, very high, medium, low, very low, or none instead of risk or no risk. Following this view, (Kangari, 1988) used fuzzy sets (i.e., low, medium, high) for linguistic expression of risks and to get the opinion from experts. Kangari developed an FST-based qualitative risk analysis model to solve the misinterpretation of risk, including a three-step risk analysis method of natural representation by FST, a fuzzy set of risks, and a linguistic approximation. (Kangari & Riggs, 1989). Sala presented a comprehensive risk management model using FST, demonstrating the use of FST in project risk assessment with risk reduction, monitoring, and control (Sala, 2015).

The fuzzy set theory provides the opportunity to use numerical and qualitative approaches to improve decision analysis in uncertainty. However, there is a need for an axial structure to encode linguistic expression in a meaningful way. Its quantitative aggregation process, choice of relationships for data, and computational methods are identified as complex problems. The validation technique of fuzzy-based decision analysis, which is important in establishing a model for practical implication, is also neglected in studies (Greaver et al., 2012). In addition, FSD-based risk assessment models have a fundamental disadvantage of not providing realistic risk assessments because sometimes avoiding complex risks makes it impossible for experts to capture all possible risk scenarios when making risk assessment judgments (Choi & Mahadevan, 2008).

## 2.1.2   EXTENDED FUZZY METHODS

### 2.1.2.1   FUZZY ALGORITHM

Fuzzy numbers can be handled by conventional arithmetic operations such as addition, subtraction, and multiplication, also known as fuzzy arithmetic. The two methods used in the fuzzy arithmetic computation are the α-cut method and the extension principle. The α-cut method leads to higher estimation due to its interval arithmetic, while the extension principle (i.e., a point-wise calculation between fuzzy input numbers and final fuzzy number as the membership degree of output points) can be used to estimate uncertainty by using different t-norms.

### 2.1.2.2   FUZZY EXPERT SYSTEM AND MAMADANI INFERENCE

Fuzzy expert systems provide an easy way to deal with situations involving fuzzy sets, both linear and uncertain properties.(Tahami et al., 2016) The results are based on expert judgment, quality assessment, causal relationships, and impact analysis (Fayek & Oduba, 2005). Any fuzzy expert systems have three basic components: fuzzy membership function, fuzzy rules, and fuzzy inference mechanism. The latter usually includes five steps: fuzzification, rule evaluation, implication, aggregation, and defuzzification (Idrus, 2011). The Mamdani inference (Jin &

Sendhoff, 2003) is an improved, more general method and a well-accepted fuzzy expert system. It is a simple "minimal operator" that reflects fuzzy "if-then" rules for deriving fuzzy inputs and fuzzy results from inference processes (Fares & Zayed, 2010; Fasanghari & Montazer, 2010).

### 2.1.3 FUZZY HYBRID METHOD

#### 2.1.3.1 FUZZY CPMREHENSIVE EVALUATION AND CLOUD MODEL

Although comprehensive fuzzy estimation (FCE) can handle uncertainty and ambiguity, the cloud model is suitable for cases involving discreteness and randomness. If project risks are vague, fuzzy and uncertain, and by nature discrete and random, and include data from domain experts, the integrated FCE and cloud model provides reliable results for assessing and prioritizing risks (Li et al., 2015). The basic components of the FCE are the risk assessment index, pair-wise comparison between risks using AHP to estimate their weight, and fuzzy weighted average and risk evaluation based on the corresponding risk score interval.

The judgment biases that can be resulted from using the maximum degree of membership are countered by generating a cloud for risk weights and risk values by a cloud model(Li et al., 2015). The cloud model is based on 'evaluation cloud' (comprising risk weights and risk values), which includes elements such as sample mean, sample variance, entropy, and excess cloud entropy (Ma et al., 2013). A 'remark cloud' is established, and decisions are made for individual risk and group risks. The basic limitations of this model are that it does not consider the impact of risks on each other within the group or beyond the group, nor any complex relationships of the risks. Therefore, if certain risks are structured in nature (with causal relationships), the model may not be suitable for risk assessment.

#### 2.1.3.2 Fuzzy fault tree and fuzzy event tree analysis

Fuzzy fault tree analysis (FTA) and fuzzy event tree analysis (ETA) are probability-based risk assessment tools that have been used for solving MCDM problems. FTA has the advantage of providing a good sketch of the root-causes of risks, making it easy to visualize and understand an event with imprecise information, although very complex project risk relationships are impossible to capture in this way (Abdelgawad & Fayek, 2011). FTA is a graphical model of some parallel events under some compound events, and a series of basic events leading to the occurrence of an unexpected event called a top event(Abdollahzadeh & Rastgoo, 2015). For FTA, the system is first defined, and a fault tree structure is constructed using an "AND" or "OR" gate, as a mediator to find the upper event from the combined effects of lower events. A minimal cut set is then measured qualitatively using Boolean algebraic analysis of a fault (basic event). Finally, a quantitative analysis is carried out to calculate the probability of occurrence of the top event. The basic drawback of this method is in obtaining an accurate estimation of the occurrence probability, although this may be overcome by adding fuzzy logic to improve the model (Abdollahzadeh & Rastgoo, 2015). This fuzzy FTA provides the fuzzy occurrence probability of the top event and offers fuzzy linguistic options for experts to assess the probability of basic events (Abdelgawad & Fayek, 2011).

The ETA is also a graphical model, where an initiating event is identified and defined, pivotal events identified and an event tree constructed. The probability of an initiating event is determined, and the binary (0, 1) probabilities of success or failure of the pivotal events are also determined. The overall probability of occurrence of each scenario is then obtained from the initiating event and pivotal events. Here, the probability of occurrence of the final event, success, and failure of an event or a scenario of occurrence is a crisp value. The accuracy of the ETA model's outcomes depends on the accuracy of the information provided. However, the use of ETA is limited, as project risks are uncertain and vague in nature. Thus, introducing fuzzy logic improves the ETA model by considering the probability of occurrence of the final event and the success/failure probability of pivotal events as a fuzzy, instead of crisp, number.(Tahami & Fakhravar, 2020b)

It also performs fuzzy arithmetic operations in ETA (Abdollahzadeh & Rastgoo, 2015).

### 2.1.3.3 FUZZY ARTIFICIAL NEURAL NETWORK (F-ANN)

The ANN is an artificial intelligence-based nonparametric model that has been used for risk analysis. ANN has the capacity to be trained from past data and be applied to generating a future outcome (Attalla & Hegazy, 2003; Loizou & French, 2012). The ANN process involves a bunch of simulated neurons (processing elements) unified in such a way that the neurons are capable of being trained (Attalla & Hegazy, 2003) ANN often produces more accurate results than other conventional methods (e.g., regression analysis) and is ideal in situations where there is lack of information regarding risks, and a complex, nonlinear or unknown relationship exists between project risks. However, the ANN model provides a single value rather than a range to define project risks (Sonmez, 2011). It has a hidden layer, which is unable to clarify the model's structure (Koo et al., 2010). In ANN, the processing elements need to be trained properly by historical data from similar projects, which mostly depends on the performance of the neurons and the availability of sufficient data (Sonmez, 2011).

Individual projects are unique in nature, and only limited risk data is available for the evaluation of project risks (Sadeghi Bazargani, 2010). As mentioned earlier, the lack of sufficient data usually means that analysts have to rely on expert judgments, and the linguistic expression is often an easy and more natural way to evaluate risks from such judgments (Abdelgawad & Fayek, 2011; Wang & Elhag, 2008). However, uncertainty exists in expert judgment due to ambiguity, vagueness, ignorance, and imprecision in understanding and evaluating the project risks (Ferdous et al., 2011) this uncertainty cannot be addressed by the traditional ANN model(Duran et al., 2012), and fuzzy logic is often used to address this application gap with ANN (Wang & Elhag, 2008). In their review work, Chan et al. (Chan et al., 2009) discovered that the combination of fuzzy and ANN would be a potential tool for uncertainty modeling in risk management.

### 2.1.3.4   FUZZY FAILURE MODE AND EFFECT ANALYSIS

Failure Mode and Effect Analysis is the process of identifying potential modes of failure in a system, evaluating the main causes, determining the impact of failures, and formulating preventive measures (Mohammadi & Tavakolan, 2013). In this system, a Risk Priority Number (RPN) for each failure mode or risk event is calculated as the product of the probability of risk occurrence (O), severity (S), and detection (D). The RPN represents the level of a particular risk, i.e., a higher value of RPN means a higher level of risk. This rating has been used in different studies to find critical risk factors (Abdelgawad & Fayek, 2011). In FMEA, "detection" is an important term exemplified as the capability of identifying the risk of not having enough working hours to take corrective action. The introduction of "detection" in prioritizing and assessing risk provides a new dimension to reach a higher level of accuracy. However, the FMEA technique assumes that S, O, and D are equally important (Zhang et al., 2012), which is not always realistic. It is also difficult to determine the precise probability of a failure event in FMEA if the data is linguistic (Xu et al., 2002) The application of fuzzy logic in FMEA can address these drawbacks in that the experts have the opportunity to assess O, S, and D in linguistic form. The fuzzy FMEA also provides an easy but efficient mechanism for modeling project risk assessment (Abdelgawad & Fayek, 2011).

### 2.1.3.5 FUZZY TOPSIS

In a fuzzy environment, the Technique for Order of Preference by Similarity to Ideal Solution (TOPSIS) is a new method that is very suitable for project selection, bid, and risk evaluation for risk assessment by multiple criteria decision analysis (MCDA) (Taylan et al., 2015). TOPSIS is a matrix method that provides a suitable and easy way of computing the weights of alternatives based on similar preferences, but it is incapable of handling any uncertainty and vagueness in expert responses (Wang & Elhag, 2008) Alternatively, the fuzzy method can handle uncertainty and vagueness, but it provides the only single value of risk as to the outcome, which is not always appropriate for reaching decisions because of some information gap. In such a case, adding TOPSIS with fuzzy can solve this problem (Hasani et al., 2012; Taylan et al., 2015). The combined fuzzy TOPSIS method can also handle both qualitative and quantitative data and provide the outcome in quantitative form for project risk assessment (Hasani et al., 2012)The method defines the fuzzy weights of risk evaluation criteria (i.e., time, cost, quality, and safety) based on expert judgment. It forms alternatives and a criteria matrix for each expert and applies a max-min rule for fuzzy inference decisions taken from the opinion of multiple experts. The matrix is then transformed into a fuzzy weighted normalized decision matrix, and positive/negative Euclidian distances and closeness coefficients are measured for all alternatives with respect to each risk. Finally, the efficiency rating for each alternative is determined for ranking the alternatives (Taylan et al., 2015).

### 2.1.3.6 FUZZY ANALYTICAL HIERARCHY (FAHP)

The AHP is a predominant MCDM technique in risk and uncertainty analysis and one of the best methods for measuring project complexity (Jato-Espino, Castillo-Lopez, Rodriguez-Hernandez, & Canteras-Jordana, 2014; L. A. Vidal et al., 2011). However, it cannot cope with inconsistencies in pairwise comparisons (Nieto-Morote & Ruz-Vila, 2011). The combination of fuzzy logic and AHP is recognized as the most influential risk management method. The fuzzy-AHP (FAHP) model can effectively measure subjective data under multiple divergent risks in a project (Zeng et al., 2007), for example, combined AHP with fuzzy logic to rank risk preferences from expert judgments. The model determines risk magnitude by aggregating the risk factor index, probability, and intensity of risk into a fuzzy decision system. However, the FAHP method has some limitations. For example, AHP does not indicate the causal relationships between the risk factors at the same level and does not consider the impact of risks in different phases of the project life-cycle, and is therefore impractical to use with a large number of risk evaluation criteria (Ebrahimnejad et al., 2012) where an enormous and tedious number of pairwise comparisons are needed. Hence, the FAHP method needs to be improved by other optimization tools.

### 2.1.3.7 FUZZY ANALYTICAL NETWORK PROCESS (FANP)

As MCDA methods, the AHP and Analytical Network Process (ANP) are frequently used for risk analysis. The AHP considers risks as independent elements of a hierarchy structure, while the risks in complex projects are highly interdependent. In contrast, the ANP captures the interdependencies and impact of various risks for risk ranking (Bu-Qammaz et al., 2009). Similar to the AHP, ANP considers pairwise comparisons of the risks, but unlike AHP, it captures all the possible causal relationships and networks between the clusters (group-risks) and among the elements (sub-risks) within a cluster. However, ANP assigns a crisp value in the pairwise comparison of risks, which is a limitation to capturing vagueness and uncertainty in risk analysis. Thus, introducing fuzzy concepts in ANP provides an advanced step in overcoming this limitation. The fuzzy-ANP (FANP) first identifies all potential risks and their interdependencies and builds a network model showing their causal relationships. It then makes pairwise comparisons between the risks using a suitable fuzzy linguistic scale or fuzzy number, tests for consistency between the data sets, aggregates the judgment matrices, calculates priority weights, computes, and limits the super-matrix, and ranks the risks based on the calculated scores. As with the AHP, however, assessing project risks based on many criteria requires an enormous number of pairwise comparisons.

### 2.1.4 FUZZY BAYESIAN BELIEF NETWORKS (F-BBN's)

The BBN can handle complex and uncertain relationships in risk networks. It is graphically defined by a directed acyclic graph (DAG) where nodes represent risks, and arrows represent the causative relationships between the risks. The arrows also denote the uncertainties inside the risk network, which are mutually inclusive under the concept of conditional probability. Using the BBN, large and complex risk networks can be easily constructed by the aggregation of sub-networks into hierarchy levels.

Two types of probabilistic data are required for any Bayesian network, the prior probability of independent risks and the conditional (effect as the influence of cause) probability of dependent risks given that of the independent risks. In the Bayesian probability theory, it is then easy to obtain the probability of a dependent risk. By placing risks in hierarchy levels, this network reduces the need for collecting a huge amount of data, as it helps in computing the probabilities of the upper level of dependent risks from the probabilities of lower-level risks (prior probability) and their probabilistic dependencies (conditional probability). Additionally, the method is a powerful tool for working with an inadequate and small number of datasets, data found from a mixture of different areas of knowledge, non-parametric and distribution-free data, and data for a highly diverse set of variables). Moreover, a BBN can easily update the probabilities in the network when new data for the variables become available. It can also deal with the prediction, deviation detection, and optimization of decision variables based on very subjective judgments

Bayesian networks assess the reliability, vulnerability, and effectiveness of a system using the probabilities of the variables and the stochastic dependencies among the variables. It computes the risk detectability and probability of false detection for assessing system reliability. In comparison with other risk assessment methods such as ANNs, MCS, CBR, and system dynamics, the BBN has a great advantage in its simplicity for use by practitioners and its accuracy with respect to the amount of data available. In BBN analysis, very precise data is required for the prior and conditional probabilities, which is difficult to obtain from large and complex projects because of the amount of uncertainty involved. There is also a lack of sufficient data for the risk assessment of complex projects, which leads to having to rely on expert opinion for data elicitation. In addition, while expert judgment is required to develop BBNs, there is limited research on how to elicit knowledge from the experts and ensure the reliability of the model. Fuzzy logic helps domain experts to express the frequency and consequences of risk linguistically, which can be transformed into a range or PERT-like three-point probability (low, medium, and high). However, fuzzy logic alone cannot express the causal relationships between the risks and is unable to conduct inverse inference (Choi & Mahadevan, 2008). Thus, a combination of fuzzy logic and BBN theory (i.e., FBBN) has a significant role to play in expediting project risk analysis in an uncertain environment

# CHAPTER 3

## DEFINITION OF RISK ASSESSMENT

Many of the project classification models have been developed in the technological arena. In 1990, Henderson and Clark examined the characteristics of innovation. They managed to differentiate the components of a product and the product "architecture," which are techniques of integrating parts of the product into the system, leading to a simple 2 x 2 model classification of the innovation, as shown in Figure below. Their studies reveal that the traditional classification of innovation is incomplete, potentially misleading, and do not take into consideration the lousy effect when small improvements to technological products are introduced. This view by (Clark, 1990) enriches the different categorizations of innovation and deepens the understanding of the relationship between innovation and organizational capability.

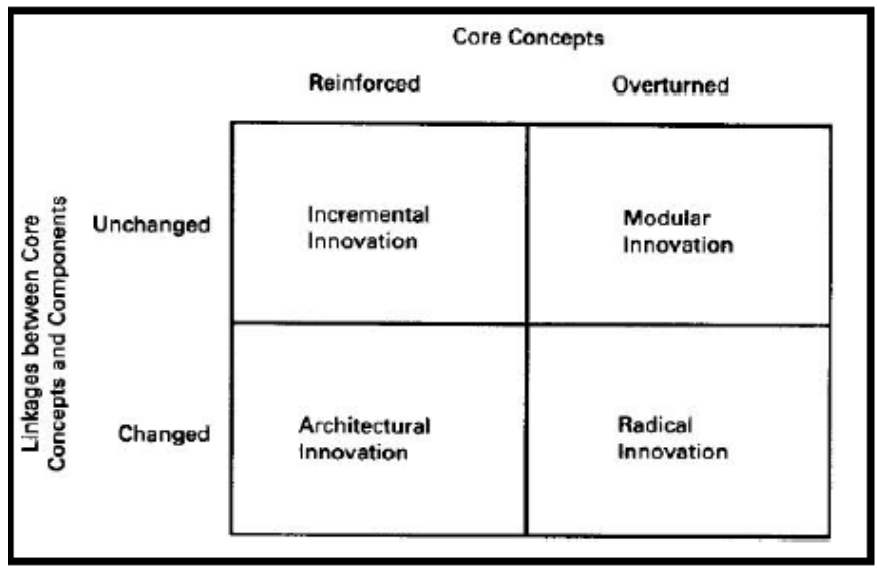

Moreover, as described by Wheelwright and (Clark, 1990), the overall project plan enables management to identify the gaps in the product development portfolio and improve the management approach of the development process. This approach also allows organizations to understand the concept of project mixing and deployment as well as overall development effort and direction to formulate a strategy for decision making.

They suggested five different levels of projects, namely derivative, breakthrough, platform, research, and development based on the level of change in the producing process and the product.

Conversely, (Turner & Cochrane, 1993) took a broader view of categorizing projects based on the level of the goals and the process of achieving the goals. The four-quadrant model is referred to as the goals-and methods matrix, which refers to was four types of projects; engineering or 'earth' projects, product development or 'water' projects, software development, or 'fire' projects and research and organizational change or 'air' projects. All four definitions of types of projects fall within the three essential aspects of product breakdown structure (PBS), organization breakdown structure (OBS), and work breakdown structure (WBS) that is needed to develop structured project governance through the selection of start-up and management techniques. Recently, expanding from the hierarchical concept of the project- facility purpose.

The model presented by Turner and Cochrane, Crawford, refers to a two-step framework that brings to mind a decision tree for developing a project classification system. It is an approach based on multiple organizational purposes and various organizational attributes to cluster projects into groups. The objectives of such classification are to develop and allocate appropriate competencies for successful implementation of projects and to prioritize projects within the investment portfolio to maximize return on investment (Keaveny & Crawford, 2006)

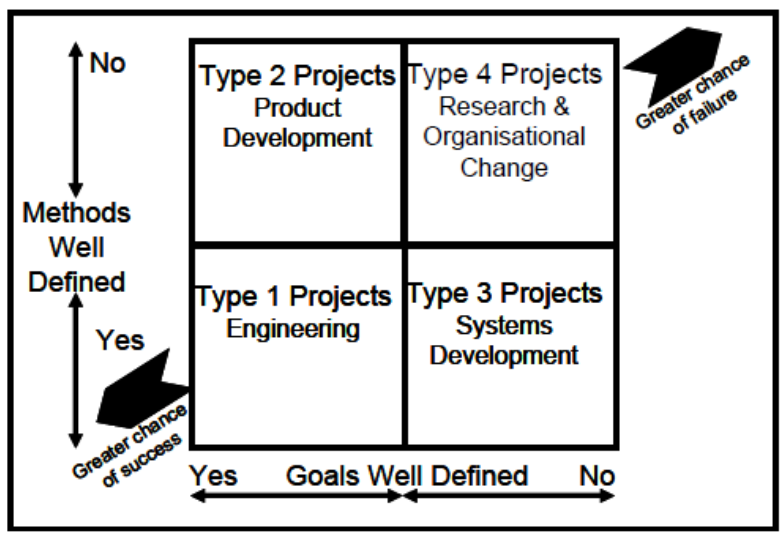

The growing importance that innovation has on organizational growth and the inexistence of any typologies that were developed into a standard, full-range and entirely accepted description of the project innovation spectrum have led Shenhar and Dvir to categorize products and innovation studies since 1992 to identify the most appropriate route for Dil (Shenhar et al., 1995). In 1996, Shenhar and Dvir successfully developed the typography theory of project management, where four types of projects were divided based on a three-dimensional structure known as the UCP Model. The four categories of projects suggested are low technical uncertainty (low-tech), medium technical uncertainty (medium-tech), high technical ambiguity (high-tech), and super-high technical uncertainty (super-high-tech) Was found to be a major factor affecting the lobes. Uncertainty, complexity, and speed are the three dimensions of the UCP model.

Because the purpose of this research was to identify and understand the best approach to risk assessment in international technology projects, we chose the Shenhar and Dvir project classification model because it is based on rigorous empirical studies and quantitative modeling. Their concept was developed to go beyond the full scope of the project classification system. Furthermore, the three dimensions suggested are found to be the factors that can represent inherent risk characteristics in projects, and their studies show the various levels of risk are identified in different types of projects.

# CHAPTER 4

# DEFINITION OF FUZZY LOGIC

In this chapter, the definition and foundational concepts of fuzzy logic, including the meaning of fuzzy set, membership function, fuzzy logical operation, and If-Then rule, are explained. This chapter also illustrates the fuzzy inference process and the features of the Mamdani-type fuzzy inference system. Meanwhile, specific procedures of Mamdani fuzzy inference are discussed in brief, and the critical concern during this process on which this research focuses is pointed out. The motivation of this study follows it. Then the last part explains the literature survey and the research organization.

## 4.1  WHAT IS FUZZY LOGIC?

In a broad sense, Fuzzy Logic is a form of soft computing method which accommodates the imprecision of the real world. In contrast to traditional, hard computing, soft computing can exploit tolerance for inaccurate, uncertainty, and partial reality, robustness, and low solution cost. In a more specific sense, Fuzzy Logic is an extension of multivalued logic whose objective is approximate reasoning rather than an exact solution. Like Binary Logic, variables may only take on truth values true and false represented by 1 and 0 respectively, the variables in Fuzzy Logic may have a truth value that levels in degree between 0 and 1. Instead of describing absolute yes or no, the truth value, or membership in Fuzzy Logic explains a matter of degree. 0 shows completely false, while one expresses entirely accurate, and any value within the range indicates the degree of truth. Furthermore, the concept of membership in Fuzzy Logic is close to human words and intuition, so a variety of applications of Fuzzy Logic and the number has increased significantly in recent years.

## 4.2  FUNDAMENTAL CONCEPTION OD FUZZY LOGIC

This research focuses on fuzzy inference, which is a primary application of fuzzy logic. The main approach of fuzzy inference is taking input variables through a mechanism that is comprised of parallel If-Then rules and fuzzy logical operations and then reach the output space. The If-Then rules are expressed directly by human words, and each of the terms is regarded as a fuzzy set. All of these fuzzy sets are required to be defined by membership functions before they are used to build If-Then rules.

## 4.3   FUZZY SET

A fuzzy set is an extension of the standard set. In the classical crisp set theory, the membership of elements complies with a binary logic; either the item belongs to the crisp set or the element does not belong to the set. In the fuzzy set theory, it may contain elements that have a degree of membership between entirely belonging to the set and completely not belonging to the set. This is because a fuzzy set does not have a crisp, clearly defined boundary, and is described by its fuzzy boundary is described by membership functions, which make the degree of membership of elements range from 0 to 1.

A brief example of the fuzzy set is showed in the last Figure. In the following fuzzy set, which describes a criterion of fuel-efficient automobiles, a model whose fuel consumption is equal or greater than 28 miles per gallon (mpg) is defined as an element in this set with a full degree of membership. In this case, an automobile with 33 mpg has a full degree of membership; in other words, completed belongs to the set. Another model with 18 mpg is far away from the criterion, so it seems they have zero degrees of membership, in other words, completed does not belong to the set. And a car with 25 mpg is fairly close to the criterion, so it is reasonable to say it has a partial degree of membership, and the feature of membership functions decides the value of membership (e.g., 0.6).

## 4.4   MEMBER FUNCTION

The member function is the defining curve that the feature of the fuzzy set by assigning to each element the corresponding membership value, or degree of membership. It maps each point in the input space to a membership value in a closed unit interval [0, 1]. The figure below shows a general membership function curve. The horizontal axis represents an input variable $x$, and the vertical axis defines the corresponding membership value $\mu(x)$ of the input variable $x$. The Support of the membership function curve explains the range where the input variable will have nonzero membership value. In this figure, $\mu(x) \neq 0$ when $x$ is any point located between point *a* and point *d*. While the Core of membership function curve interprets the range where the input variable *x* will have the full degree of membership ( $\mu(x) = 1$ ), in other words, the arbitrary point within the interval [*b, c*] completely belongs to a fuzzy set which is defined by this membership function.

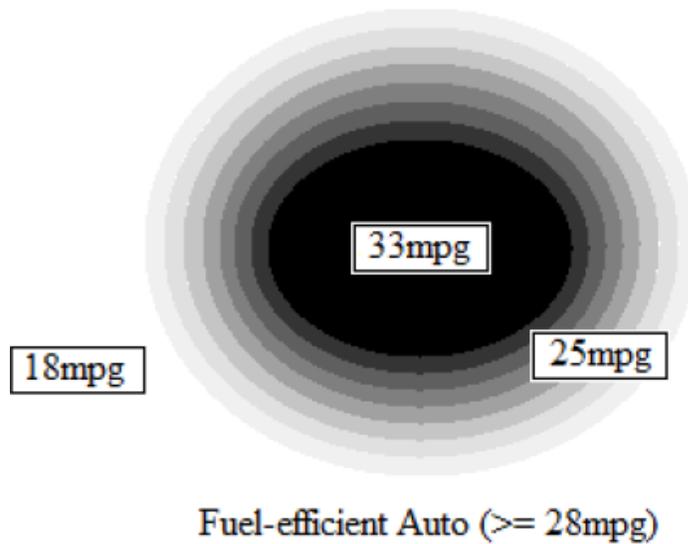

Fuel-efficient Auto (>= 28mpg)

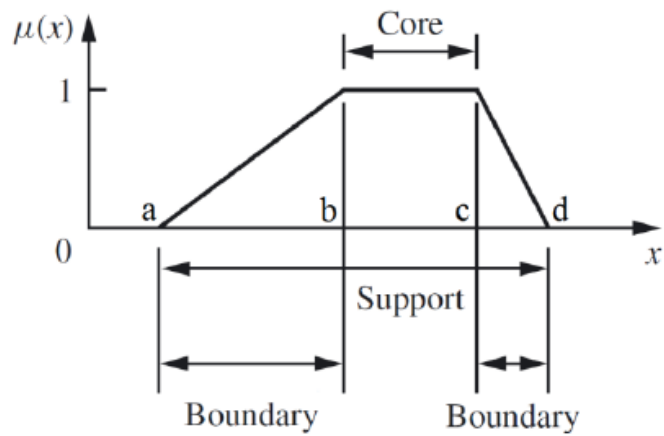

Generally, there are five common shapes of membership function: Triangle MF, Trapezoidal MF, Gaussian MF, Generalized Bell MF, and Sigmoidal MF. Regardless of the shape, a single MF may only define one fuzzy set. Usually, more than one MF is used to describe a single input variable. Taking the fuel consumption of automobiles, for instance, a three-level fuzzy system with fuzzy sets 'Low', 'Medium,' and 'High' is applicable to represent the whole situation.

## 4.5   LOGIC OPERATION

Because the standard binary logic is a particular case of fuzzy logic where the membership values are always 1 (completely true) or 0 (completely false), fuzzy logic must hold the consistent logical operations as the standard logical operations. The most foundational logical operations are AND, OR, and NOT. Unlike the standard logical process, the operands $A$ and $B$ are membership values within the interval [0, 1]. In fuzzy logical operations, valid AND is expressed by function *min*, so statement $A$ AND $B$ is equal to *min* ($A$, $B$). Reasonable OR is defined by function *max*; thus, $A$ OR $B$ becomes equivalent to *max* ($A$, $B$). And valid NOT make operation NOT $A$ become the operation $1 - A$.

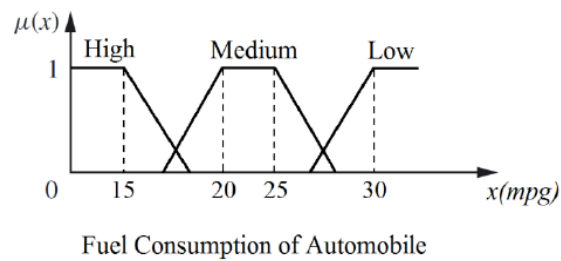

Fuel Consumption of Automobile

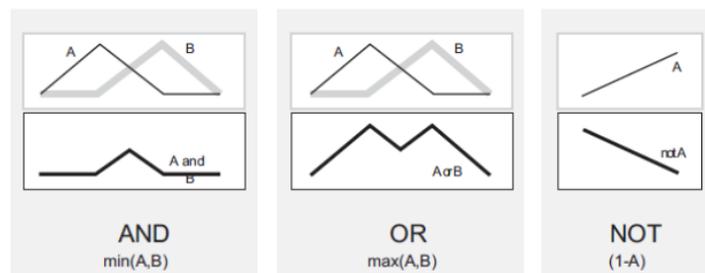

## 4.6   IF-THEN RULES

In the fuzzy inference process, parallel If-Then rules form the deducing mechanism, which indicates how to project input variables onto output space. A single fuzzy If-Then rule follows the form If $x$ is $A$, Then $y$ is $B$ The first If-part is called the antecedent, where $x$ is the input variable. The rest Then-part is called the consequent, and $y$ is the output variable. The reason why If-Then additional statements are universally applicable is that both $A$ and $B$ are linguistic values or adjectives in most cases, and this form of conditional statement works the concordant way with human judgment. For example, an appropriate If-Then rule might be "If *material hardness* is *hard*, Then *cutting speed* is s*low*." $A$ can be regarded as a fuzzy set and defined by a specific membership function. $B$ can be either a fuzzy set or a polynomial concerning input $x$ depending on a specific fuzzy inference method. In the antecedent, the If part is aimed at working out the membership value of input variable $x$ corresponding to fuzzy set $A$. While in the consequent, the Then-part assigns a crisp value back to the output variable $y$.

## 4.7   FUZZY INFERENCE AND MAMDANI- TYPE FUZZY INFERENCE

Fuzzy inference is the process of mapping the given input variables to an output space via a fuzzy logic-based deducing mechanism, which is comprised of If-Then rules, membership functions, and fuzzy logical operations. Because the form of If-Then rule fits in human reasoning, and fuzzy logic approximates to people's linguistic habits, this inference process 9 projecting crisp quantities onto human language and promptly yielding a precise value, as a result, is widely adopted.

Generally, three types of fuzzy inference methods are proposed in the literature: *Mamdani* fuzzy inference, *Sugeno* fuzzy inference, and *Tsukamoto* fuzzy inference. All of these three methods can be divided into two processes. The first process is fuzzifying the crisp values of input variables into membership values according to appropriate fuzzy sets, and these three methods are the same in this process. Although difference exists in the second method, the effects od all rules are combined into a single, reliable output value.

In Mamdani inference, the consequent of If-Then rule is defined by the fuzzy set. The fuzzy output set of each rule will be reshaped by a matching number, and defuzzification is required after aggregating all of these reshaped fuzzy sets. But in Sugeno inference, the consequent of If-Then rule is explained by a polynomial with respect to input variables; thus, the output of each rule is a single number. Then a weighting mechanism is implemented to work out the final crisp output. Although Sugeno inference avoids the complex defuzzification, the work of determining the parameters of polynomials is inefficient and less straightforward than defining the fuzzy output sets for Mamdani inference. Thus Mamdani inference is more popular, and this research only focuses on the Mamdani inference method. Tsukamoto inference seems like a combination of Mamdani and Sugeno method, but it is even less transparent than these two models.

## 4.8   DEFUZZIFICATION

The last step of the fuzzy inference process is defuzzification, through which the combined fuzzy set from the aggregation process will output a single scalar quantity. As the name implies, defuzzification is the opposite operation of fuzzification. Since, in the first procedure, the crisp values of input variables are fuzzified into the degree of membership concerning fuzzy sets, the last process extracts a precise quantity out of the range of fuzzy set to the output variable. Among the many defuzzification methods that have been proposed in the literature, the *Centroid Method* (also called the *center of area* or *center of gravity*), which is the most popular and appealing of all the defuzzification methods, is the only adopted method in this research. The algebraic expression gives it

$$z_{COA} = \frac{\int_z \mu_A(z) \cdot z \, dz}{\int_z \mu_A(z)}$$

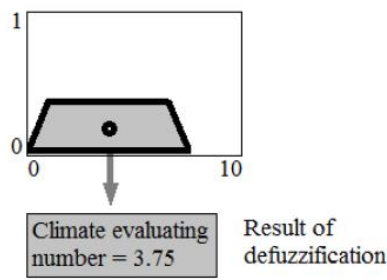

# CHAPTER 5

# RISK ASSESSMENT FRAMEWORK
# BASED ON FUZZY LOGIC

## 5.1 RISK ASSESSMENT AND DECISION-MAKING

A Structural risk assessment and decision-making platform in a fuzzy logic system will provide consistency when analyzing risks with limited data and knowledge. It allows people to focus on the base of risk assessment, which includes the cause-and-effect relationship between critical factors and the exposure for each risk. Direct input for the likelihood and possibility severity of a risk event, it encourages human reasoning to end in a consistent and well-documented way from facts and knowledge.

The graph below shows a model risk assessment process based on the fuzzy logic system. It is a downstream system that starts with each risk. Risk exposure is then integrated at both business exposure and enterprise levels to identify greater risks

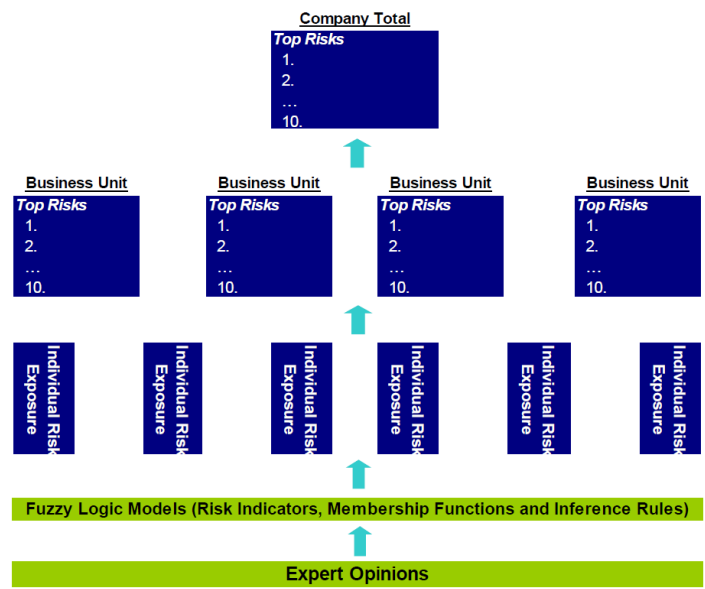

The same should be observed when evaluating exposure to each risk. The estimated amount of loss under a potential candidate for extreme cases. If we can simulate the distribution of loss using the fuzzy logic model given by the distribution of independent variables, this measure can be as high as the 99.5th percentile of the loss distribution (1-in-200 year event).

By using the loss amount as the output variable, risks can be estimated as a result of a numerical value that measures the level of risk exposure(defuzzification). It is the same as a ranking based on the level of the fact that the risk exposure is high. The figure below illustrates the classification of risks based on the estimated amount of loss under extreme events. The amount of loss resulting from confusion can be determined using the fuzzy logic model. The amount of loss may be the output variable in the fuzzy logic model. In the extreme case, the value of the input variables is paid into the model to derive the estimated loss amount for a particular risk. An alternative concept is the use of simulation, as illustrated in Figure below. The distribution of the loss amount can be simulated, and the value at a given percentage can be used to represent risk exposure

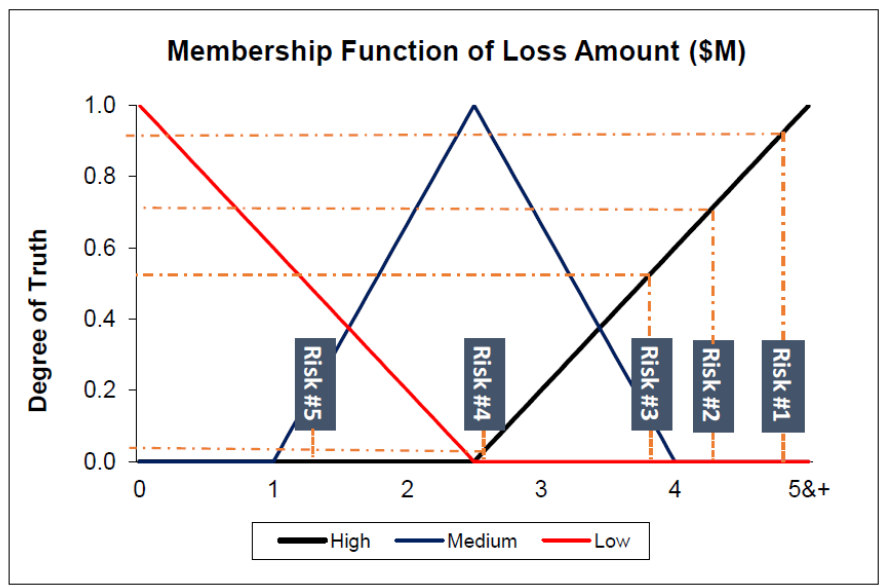

Also, to helping identify the top risks, fuzzy logic models may contain information about the risk exposure or factors that have a huge impact on it. This may provide clues to the direction of risk methods. The cost of risk-hedging can be added as an additional output variable in the fuzzy model.

This will help management decide which risks are mitigated and the most cost-effective approach to doing so. The debate so far suggests that all the experts share the same view of risks. Given the different levels of understanding and experience of experts, this is unlikely to be true in actual practice. Therefore, it is necessary to gather diverse opinions. There are many approaches to mobilization.

1. Adjust inference rules and the membership functions to integrate different ideas. The example is shown in Figure below, the weighted average of the member functions provided by Expert A and Expert B can be used as a combined members function for the high-level fuzzy set. Weights can be determined or changed based on each expert's experience, knowledge of the problem being investigated, the accuracy of past assessments, and confidence in his/her opinion.

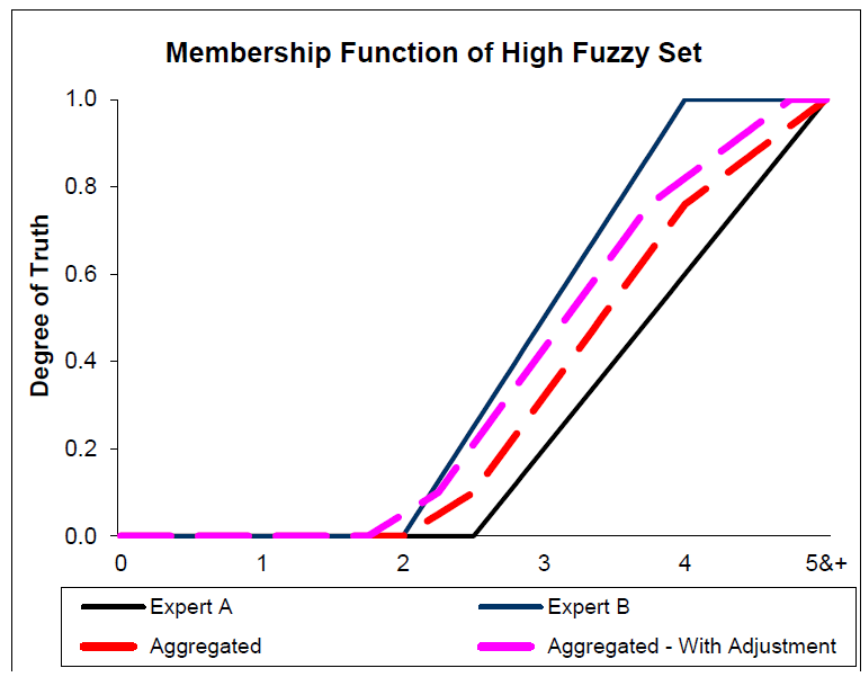

It is also possible to have differing opinions about the inference rules. If the difference is not very large, the adjustment in the membership function can combine that difference. Suppose there are two assumed rules:

*Expert I: If X is low, then Y is low.*

*Expert II: If X is not low (medium or high), then Y is high.*

The integral member function of a high fuzzy set can be shifted to the left. When we are changing the membership function of the high fuzzy set, this part shows the inference rule in this fuzzy system that *If X is medium, then Y is high*. Only one inference rule should be included in the fuzzy logic model.

*If X is low, then Y is low*.

However, if there are counter opinions about the inference rules, then we need to understand the reasoning behind each concept. Experts can edit their opinions after learning from the opposite side. If there are more conflicting ideas at the end of the discussion, then both inference rules may be removed completely from the model since the disagreement may indicate a lack of knowledge and a low level of credibility.

2. Each expert can have his/her own fuzzy logic model with unique membership functions and inference rules. The combined risk assessment result is the weighted average of the results generated from the different each model.

Unlike the first approach of adjusting the model inputs, the second fits the model outputs by melding them all together

3. A particular case of the second approach is to assign equal weight to all views, which exist in the literature about fuzzy logic models. This is usually used when there are a few experts, and the goal is to rank based on the level of risk.

# CHAPTER 6

## CONCLUSION

Fuzzy logic models, as a complement to probability models, can be used to estimate the risks of inadequate data and incomplete knowledge. Fuzzy logic provides a framework where human rational and inaccurate data may contribute to risk analysis. The scope of possible applications for fuzzy logic systems is extensive. Many risks are out of control, not well understood or known, as evidenced by the growing list of risks involved.

It is possible to analyze multiple risks that are not well understood consistently. Exposure to each risk can be assessed and ranked. Identify and manage key risks. Resources can be used to monitor and mitigate these key risks with greater exposure. The inference rules in a fuzzy logic model may not only help identify the cause of a particular risk but also help design effective mitigation plans.

Fuzzy logic systems enable us to develop knowledge of risks in two ways.

1. The systems allow risk managers and subject matter experts to focus on cause-and-effect relationships based on multiple risk assumptions and their knowledge of the system.

2. Risk assessment results related to the risk decision-making process and decision can be re-entered into the system to enrich the fuzzy sets and rules.

Fuzzy logic models can be used with other risk models such as decision trees and artificial neural networks to model complex risk issues like policyholder behaviors.